\documentclass[10pt,twocolumn,letterpaper]{article}

\usepackage{cvpr}
\usepackage{times}
\usepackage{epsfig}
\usepackage{graphicx}
\usepackage{amsmath}
\usepackage{amssymb}

\usepackage{subcaption}
\usepackage{pdfpages}

\usepackage[pagebackref=true,breaklinks=true,letterpaper=true,colorlinks,bookmarks=false]{hyperref}

\cvprfinalcopy % *** Uncomment this line for the final submission

\def\cvprPaperID{4226} % *** Enter the CVPR Paper ID here

\ifcvprfinal\fi
\begin{document}

%%%%%%%%% TITLE
%\title{Image Inpainting using PGGAN: Patch-Global Generative Adversarial Networks}
\title{Patch-Based Image Inpainting with Generative Adversarial Networks}

\author{Ugur Demir\\
Istanbul Technical University\\
%Institution1 address\\
{\tt\small ugurdemir@itu.edu.tr}
% For a paper whose authors are all at the same institution,
% omit the following lines up until the closing ``}''.
% Additional authors and addresses can be added with ``\and'',
% just like the second author.
% To save space, use either the email address or home page, not both
\and
Gozde Unal\\
Istanbul Technical University\\
%First line of institution2 address\\
{\tt\small unalgo@itu.edu.tr}
}

\makeatletter
\let\@oldmaketitle\@maketitle% Store \@maketitle
\renewcommand{\@maketitle}{\@oldmaketitle% Update \@maketitle to insert...
  \includegraphics[width=\linewidth]
    {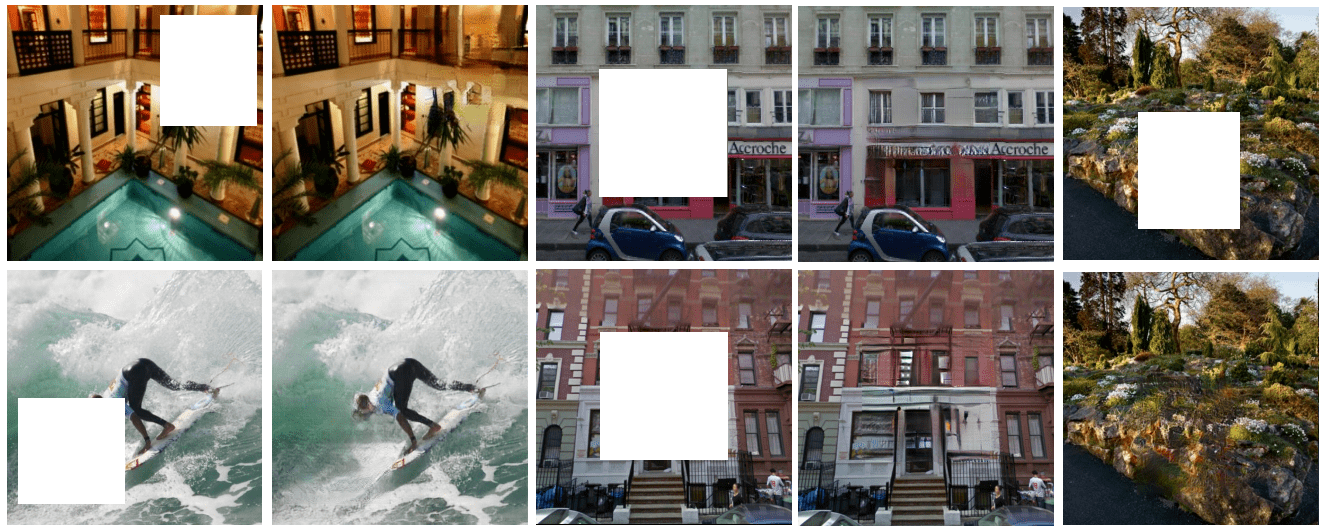}
   % \text{Figure 0: Inpainting results of the proposed PGGAN: patchGAN-globalGAN.}
    \bigskip
    }% ... an image
\makeatother

\maketitle
%\thispagestyle{empty}

%%%%%%%%% ABSTRACT
\begin{abstract}
\vspace*{-0.2cm}
Area of image inpainting over relatively large missing regions recently advanced substantially through adaptation of dedicated deep neural networks. However, current network solutions still introduce undesired artifacts and noise to the repaired regions. We present an image inpainting method that is based on the celebrated generative adversarial network (GAN) framework. The proposed PGGAN method includes a discriminator network that combines a global GAN (G-GAN) architecture with a patchGAN approach. PGGAN first shares network layers between G-GAN and patchGAN, then splits paths to produce two adversarial losses that feed the generator network in order to capture both local continuity of image texture and pervasive global features in images. The proposed framework is evaluated extensively, and the results including comparison to recent state-of-the-art demonstrate that it achieves considerable improvements on both visual and quantitative evaluations.
\end{abstract}

%%%%%%%%% BODY TEXT
\section{Introduction}
%-------------------------------------------------------------------------

%\begin{figure}[t]
%\begin{center}
%\fbox{\rule{0pt}{3in} \rule{0.9\linewidth}{0pt}}
   %\includegraphics[width=0.8\linewidth]{egfigure.eps}
%\end{center}
%   \caption{Showcase results for the first impression. placeholder text}
%\label{fig:teaser}
%\end{figure}

%-------------------------------------------------------------------------
Image inpainting is a widely used reconstruction technique by advanced photo and video editing applications for repairing damaged images or refilling the missing parts. The aim of the inpainting can be stated as reconstruction of an image without introducing noticeable changes. Although fixing small deteriorations are relatively simple, filling large holes or removing an object from the scene are still challenging due to huge variabilities and complexity in the high dimensional image texture space. We propose a neural network model and a training framework that completes the large blanks in the images. As the damaged area(s) take up large space, hence the loss of information is considerable, the CNN model needs to deal with both local and global harmony and conformity to produce realistic outputs. 

Recent advances in generative models show that deep neural networks can synthesize realistic looking images remarkably, in applications such as super-resolution \cite{vdsr, srgan,srcnn}, deblurring \cite{dblur}, denoising \cite{dncnn} and inpainting \cite{context_encoder, high_res_mc, glgan, face_comp}. One of the essential questions about realistic texture synthesis is: how can we measure "realism" or "naturalness"? One needs to formulate a yet inexistent formulation or an algorithm that determines precisely whether an image is real or artificially constructed. Primitive objective functions like Euclidean Distance assist in measuring and comparing information on the general structure of the images, however, they tend to converge to the mean of possible intensity values that cause blurry outputs. In order to solve this challenging problem, Goodfellow \etal proposed Generative Adversarial Networks (GAN) \cite{gan}, which is a synthesis model trained based on a comparison of real images with generated outputs. Additionally, a discriminative network is included to classify whether an image comes from a real distribution or a generator network output. During the training, the generative network is scored by an adversarial loss that is calculated by the discriminator network. 

Grading a whole image as real or fake can be employed for small images \cite{context_encoder}, however  high resolution synthesis needs to pay more attention to local details along with the global structure \cite{high_res_mc, glgan, face_comp}. Isola \etal introduced the PatchGAN that reformulates the discriminator in the GAN setting to evaluate the local patches from the input \cite{pix2pix}. This work showed that PatchGAN improves the quality of the generated images, however it is not yet explored for image inpainting. We design a new discriminator that aggregates the local and global information by combining the global GAN (G-GAN) and PatchGAN approaches for that purpose.

In this paper, we propose an image inpainting architecture with the following contributions:
\begin{itemize}
\item Combination of PatchGAN and G-GAN that first shares network layers, later uses split paths with two separate adversarial losses in order to capture both local continuity and holistic features in images;
\item Addition of dilated  and interpolated convolutions to ResNet \cite{perceptual} in an overall end-to-end training network created for high-resolution image inpainting;
\item Analysis of different network components through ablation studies;
\item A detailed comparison to latest state-of-the-art inpainting methods.
\end{itemize}

%-------------------------------------------------------------------------

\begin{figure}[t]
\begin{center}
	%\fbox{\rule{0pt}{2in} \rule{0.9\linewidth}{0pt}}
	\includegraphics[width=0.8\linewidth]{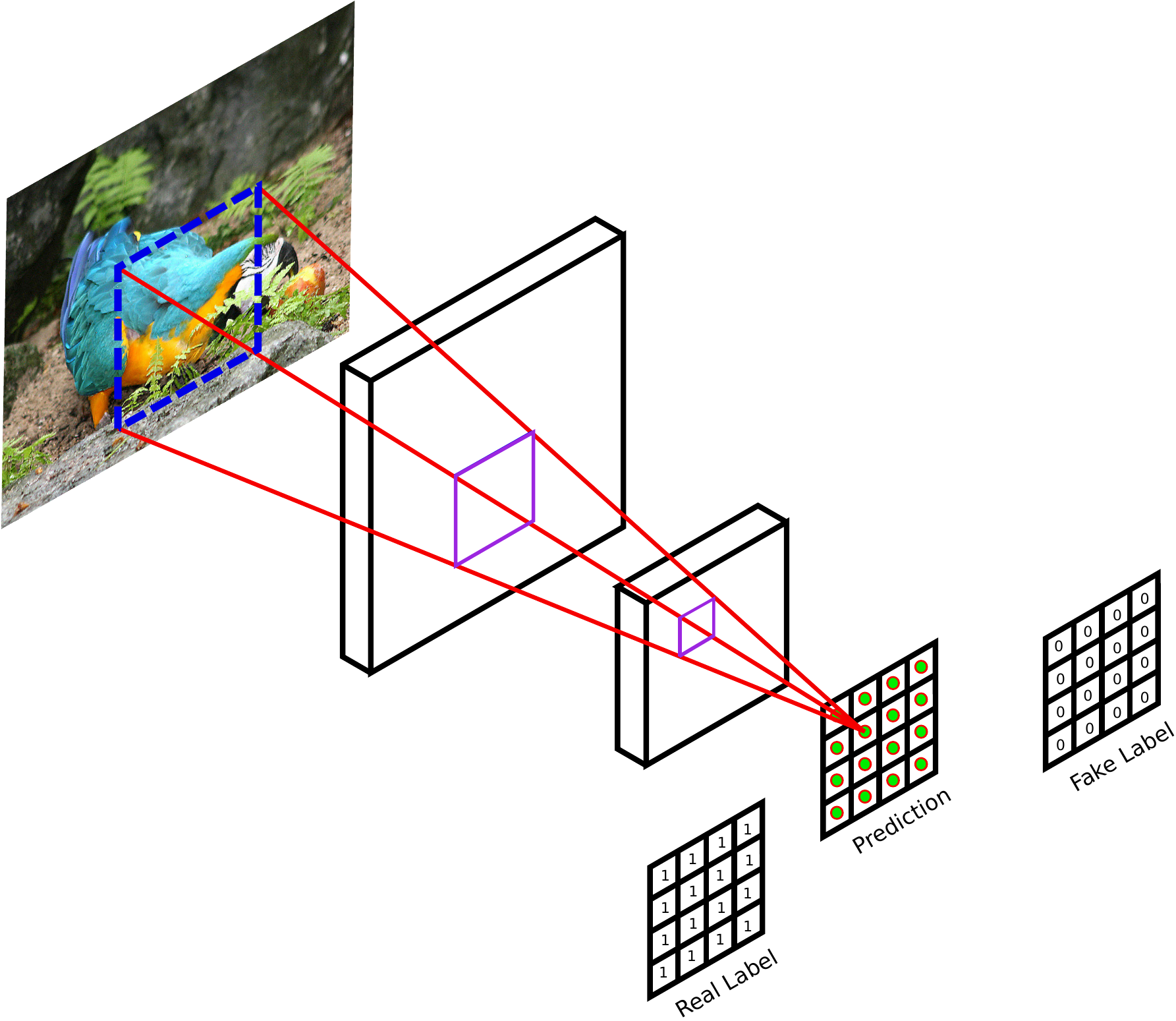}
\end{center}
   \caption{PatchGAN discriminator. Each value of the output matrix represents the probability of whether the corresponding image patch is real or it is artificially generated.}
\label{fig:patchgan}
\end{figure}

%-------------------------------------------------------------------------

\begin{figure*}[t]
\begin{center}
%\fbox{\rule{0pt}{2in} \rule{0.9\linewidth}{0pt}}
   \includegraphics[width=1.0\linewidth]{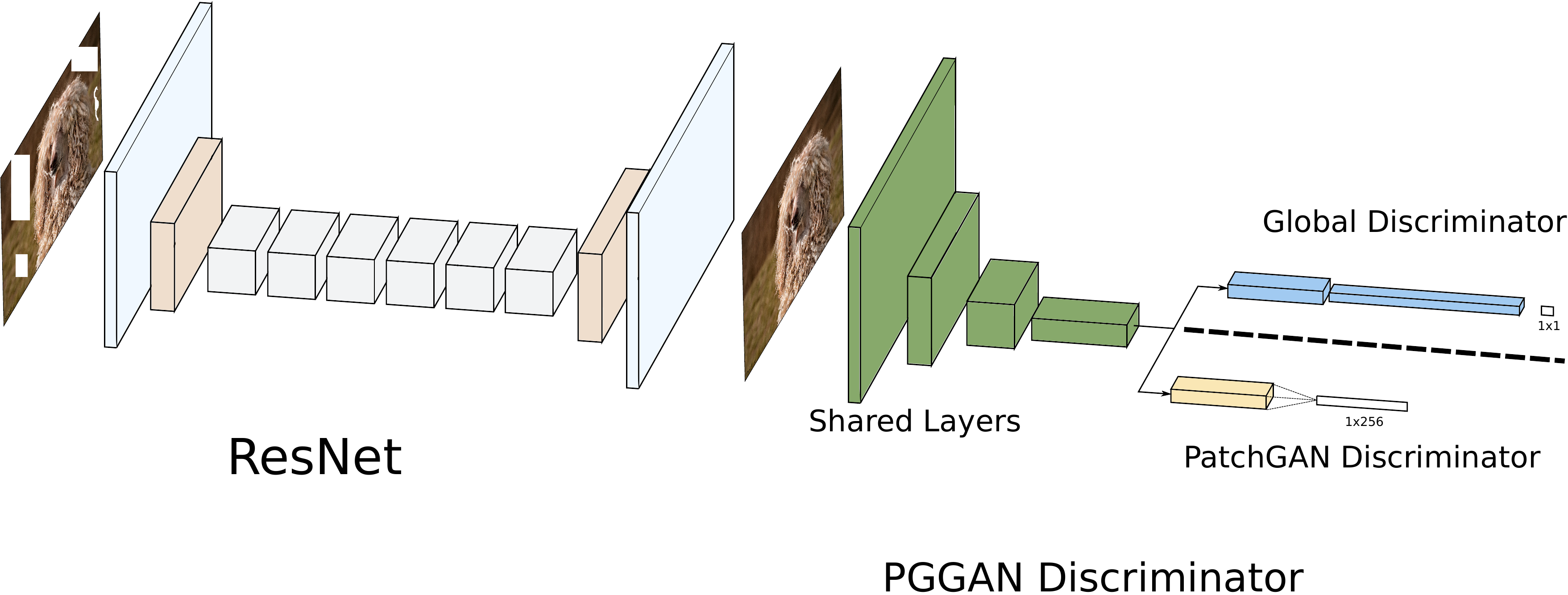}
\end{center}
   \caption{Generative ResNet architecture and PGGAN discriminator which is formed by combining PatchGAN and G-GAN.}
\label{fig:g_arch}
\end{figure*}

%-------------------------------------------------------------------------
\section{Related works}
The idea of \textbf{AutoEncoders} (AE) dominated the generative modeling literature in the last decade. Theoretical developments in connecting probabilistic inference with efficient approximate optimization as in Variational AutoEncoders \cite{vae} and the intuitive expansion of AEs to Denoising Autoencoders (DAE) \cite{dae} constitute building blocks of image synthesis models both in terms of theory and neural network (NN) implementations. Particularly, the design of NN architectures has a crucial effect on texture generation as it shapes the information flow through the layers as desired. The AE framework transforms the input image to an abstract representation, then recover the image from learnt features. To improve gradient flow in backpropagations, skip connections are added to improve synthesis quality in \cite{unet}. Residual connections \cite{resnet, identity_map, wide_res, inception, aggregated_res} that enhance the gradient flow are also adapted to generative models \cite{perceptual, pix2pix, dncnn, ct_reconstruct, darn}. Apart from the architectural design, recently introduced components as batch normalization \cite{batch_norm}, instance normalization \cite{instance_norm}, dilated convolution \cite{dilated_conv} and interpolated convolution \cite{iconv} produce promising effects on the results of image generation process \cite{perceptual, unet, srgan, vdsr, glgan}.

\textbf{Adversarial training} has become a vital step for texture generator Convolutional Neural Networks (CNNs). It provides substantial gradients to drive the generative networks toward producing more realistic images without any human supervision. However, it suffers from unstable discriminator behavior during training which frustrates the generator convergence. Furthermore, the GAN considers images holistically and focuses solely on the realistic image generation rather than generation of an image patch well-matched to the global image. That property of GAN is incompatible with the original goal of the inpainting. Numerous GAN-like architectures have been proposed during the last years to solve those issues to some degree \cite{ebgan, ppgan, categor_gan, pyramid_gan, pix2pix}.

Recently proposed PatchGAN \cite{pix2pix, mcmc} provides a simple framework that can be adapted to various image generation problems. Instead of grading the whole image, it slides a window over the input and produces a score  that indicates whether the patch is real or fake. As the local continuity is preserved, a generative network can reveal more detail from the available context as illustrated in the cover figure which presents some results of the proposed technique. To our knowledge, our work is the first to accommodate PatchGAN approach to work with the inpainting problem.

\textbf{Inpainting}: Early inpainting studies, which worked on a single image, \cite{bertalmio_inpaint, crimsi, graph_cut, patch_match} typically created solutions through filling the missing region with texture from similar or closest image areas, hence they suffered from the lack of global structural information.

A pioneering study that incorporated CNNs into the inpainting is proposed by Pathak \etal \cite{context_encoder}. They developed Context-Encoder (CE) architecture and applied adversarial training \cite{gan} to learn features while regressing the missing part of the images. Although the CE had shown promising results, inadequate representation generation skills of an AutoEncoder network in the CE led to substantial amount of implausible results as well.

An importance-weighted context loss that considers closeness to the corrupted region is utilized in \cite{semantic}. In Yang \etal \cite{high_res_mc}, a CE-like network is trained with an adversarial and a Euclidean loss to obtain the global structure of the input. Then, the style transfer method of \cite{mcmc} is used, which forces features of the small patches from the masked area to be close to those of the undamaged region to improve texture details.

Two recent studies on arbitrary region completion \cite{face_comp, glgan} add a new discriminator network that considers only the filled region to emphasize the adversarial loss on top of the global GAN discriminator (G-GAN). This additional network, which is called the local discriminator (L-GAN), facilitates exposing the local structural details. Although those works have shown prominent results for the large hole filling problem, their main drawback is the LGAN's emphasis on conditioning to the location of the mask. It is observed that this leads to disharmony between the masked area where the LGAN is interested in and the uncorrupted texture in the unmasked area. The same  problem is indicated in \cite{glgan} and solved by applying post-processing methods to the synthesized image. In \cite{face_comp}, LGAN pushes the generative network to produce independent textures that are incompatible with the whole image semantics. This problem is solved by adding an extension network that corrects the imperfections. Our proposed method on the other hand explores every possible local region as well as dependencies among them to exploit local information to the fullest degree.

\section{Proposed Method}
We introduce a generative CNN model and a training procedure for the arbitrary and large hole filling problem. The generator network takes the corrupted image and tries to reconstruct the repaired image.  We utilized the ResNet \cite{perceptual} architecture as our generator model with a few alterations. 
During the training, we employ the adversarial loss to obtain realistic looking outputs. The key point of our work is the following: we design a novel discriminator network that combines G-GAN structure with PatchGAN approach which we call PGGAN. The proposed network architecture is shown in Figure \ref{fig:g_arch}.

%-------------------------------------------------------------------------

\begin{figure}[h]
\begin{center}
	%\fbox{\rule{0pt}{2in} \rule{0.9\linewidth}{0pt}}
	\includegraphics[width=0.8\linewidth]{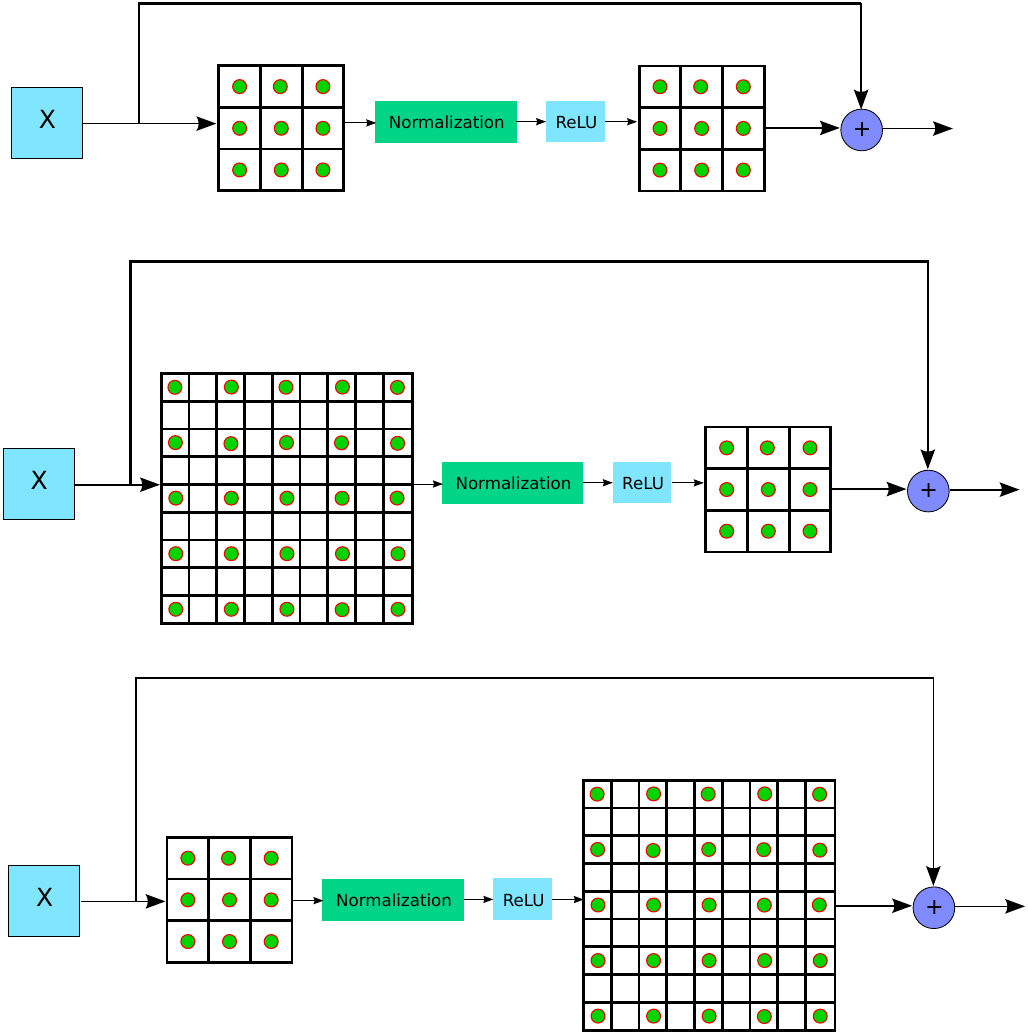}
\end{center}
   \caption{Residual block types. a: standard residual block. b: dilated convolution is placed first. c: Dilated convolution is placed second.}
\label{fig:res_blocks}
\end{figure}

%-------------------------------------------------------------------------

\subsection{Generator network}

The generative ResNet that we compose consists of down-sampling, residual blocks and up-sampling parts using the architectural guidelines introduced in \cite{perceptual}. Down-sampling layers are implemented by using strided convolutions without pooling layers. Residual blocks do not change the width or height of the activation maps.  Since our network performs completion operation in an end-to-end manner, the output must have the same dimension with the input. Thus, in the configuration of all our experiments, the number of down-sampling and up-sampling layers are selected as equal. 

Receptive field sizes, which dictate dependency between distant regions, have a critical effect on texture generation. If the amount of sub-sampling is raised to increase the receptive field, the up-sampling part of the generator network will be faced with a more difficult problem that typically leads to low quality or blurry outputs. The \textbf{dilated convolution} operation is utilized in \cite{dilated_conv} in order to increase the receptive field size without applying sub-sampling or adding excessive amount of convolution layers. Dilated convolution spreads out the convolution weights to over a wider area to expand the receptive field size significantly without increasing the number of parameters. This was first used by \cite{glgan} for inpainting. We also investigate the effect of the dilated convolution for texture synthesis problem.  Three different residual block types are used in our experiments as shown in the Figure \ref{fig:res_blocks}. First residual block which is called type-a contains only two standard convolutions, normalization, activation and a residual connection. Other types introduce dilated convolution. Type-b block places dilation before the normalization layer and type-c block uses dilation after the activation layer. While dilation is used in our network, dilation parameter is increased by a factor of two in each residual block starting from one. 

\textbf{Interpolated convolution} is proposed by Odena \etal \cite{iconv} to overcome the well-known checkerboard artifacts during the up-sampling operation caused by the transposed convolution (also known as deconvolution). Instead of learning a direct mapping from a low resolution feature map to high resolution, the input is resized to the desired size and then the convolution operation is applied. Figure \ref{fig:upsample_cmp} shows how the interpolated convolution affects the image synthesis elegantly.

\subsection{Discriminator network}
Discriminator network D takes the generated and real images and aims to distinguish them while the generator network G makes an effort to fool it. As long as D successfully classifies its input, G benefits from the gradient provided by the D network via its adversarial loss.

We achieve our goal of obtaining an objective value that measures the quality of the image as a whole as well as the consistency in local details through our PGGAN approach depicted in Figure \ref{fig:g_arch}. Rather than training two separate networks simultaneously, we design a weight sharing architecture at the first few layers so that they learn common low level visual features. After a certain layer, they are split into two pathways.  The first path ends up with a binary output which decides whether the whole image is real or not. The second path evaluates the local texture details similar to the PatchGAN. Fully connected layers are added at the end of the second path of our discriminator network to reveal full dependency across the local patches. The overall architecture hence provides an objective evaluation of the naturalness of the whole image as well as the coherence of the local texture.

%-------------------------------------------------------------------------

\subsection{Objective function}
At the training stage, we use a combination of three loss functions. They are optimized jointly via backpropagation using Adam optimizer \cite{adam}. We describe each loss function briefly as follows.

{\bf Reconstruction loss} computes the pixel-wise L1 distance between the synthesized image and the ground truth. Even though it forces the network to produce a blurry output, it guides the network to roughly predict texture colors and low frequency details. It is defined as:
\begin{equation}
\mathcal{L}_{rec} = \frac{1}{N}\sum_{n=1}^N\frac{1}{WHC}||y-x||_1
\end{equation}
where $N$ is the number of samples, $x$ is the ground truth, $y$ is the generated output image, $W$, $H$, $C$ are width, height, and channel size of the images, respectively. 

{\bf Adversarial loss} is computed by the both paths of PGGAN discriminator network D that is introduced in the training phase. Generator G and D are trained simultaneously by solving $\arg \min_G \max_D L_{GAN}(G,D)$:
\begin{eqnarray}
\label{eq:adv_minmax}
\mathcal{L}_{GAN}(G,D) &=& \mathbb{E}_{x\sim p(x)}[\log D(x)] \nonumber \\
             &+& \mathbb{E}_{y\sim p_{G}(\tilde{x})}[\log (1 - D(G(\tilde{x})))]
\end{eqnarray}
where $\tilde{x}$ is the corrupted image.

\textbf{Joint loss} function defines the objective used in the training phase. Each component of the loss function is governed by a coefficient $\lambda$:
\begin{equation}
\label{eq:loss_joint}
\mathcal{L} = \lambda_{1} \mathcal{L}_{rec} + \lambda_{2} \mathcal{L}_{g\_adv} + \lambda_{3} \mathcal{L}_{p\_adv}
\end{equation}
where $\mathcal{L}_{g\_adv}$ and $\mathcal{L}_{p\_adv}$ refer to $\mathcal{L}_{GAN}$ in Equation \ref{eq:adv_minmax} corresponding to two output paths of the PGGAN (see Figure \ref{fig:res_blocks}). We update the generator parameters by joint loss $\mathcal{L}$,  unshared G-GAN layers by $\mathcal{L}_{g\_adv}$, unshared P-GAN layers by $\mathcal{L}_{p\_adv}$ and shared layers by $\mathcal{L}_{g\_adv} + \mathcal{L}_{p\_adv}$.

%-------------------------------------------------------------------------

\section{Results}
\label{sec:exp}
In this section, we evaluate the performance of our method and compare PGGAN with the recent inpainting methods through ablation studies, quantitative measurements, perceptual scores and visual evaluations.

\subsection{Datasets}
\textbf{Paris Street View} \cite{psv} has 14900 training images and 100 test images which is collected from Paris. Comparisons and our ablation study are mostly performed on this dataset.

\textbf{Google Street View} \cite{crcv} consist of 62058 high quality images. It is divided into 10 parts. We use the first and tenth parts as the testing set, the ninth part for validation, and the rest of the parts are included in the training set. In this way, 46200 images are used for training.

\textbf{Places} \cite{places} is one of the largest dataset for visual tasks that has nearly 8 million training images. Since there is considerable amount of data in the set, it is helpful for testing generalizability of out networks.

\subsection{Training details and implementation}
All of the experimental setup is implemented using Pytorch \footnote{http://pytorch.org/} with GPU support. Our networks are trained separately on four NVIDIA\textsuperscript{TM} Tesla P100 and a K40 graphic cards.

In order to obtain comparable results from our generative ResNet implementation, we use 3 subsampling blocks when type-a blocks are used. If dilated convolution is used in the residual blocks, subsampling is set to two since dilation parameter makes it possible to reach wider regions without subsampling.

While training our networks with PGGAN discriminator, we set $\lambda_{1} = 0.995$, $\lambda_{2} = 0.0025$ and $\lambda_{3} = 0.0025$ in Equation \ref{eq:loss_joint}.

%-------------------------------------------------------------------------
\begin{figure}[h]
   \centering
\begin{tabular}{c@{\hskip 0.04in}c@{\hskip 0.04in}c@{\hskip 0.04in}c}

\rotatebox[origin=l]{90}{G-GAN}&
\includegraphics[width=0.3\linewidth]{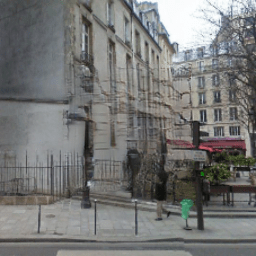}&
\includegraphics[width=0.3\linewidth]{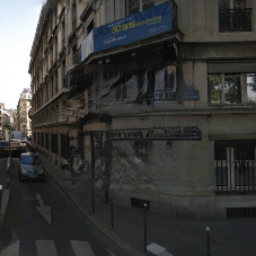}&
\includegraphics[width=0.3\linewidth]{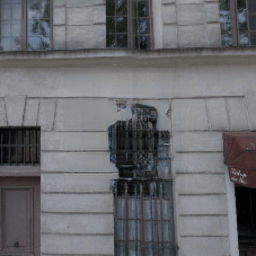}\\

\rotatebox[origin=l]{90}{PatchGAN}&
\includegraphics[width=0.3\linewidth]{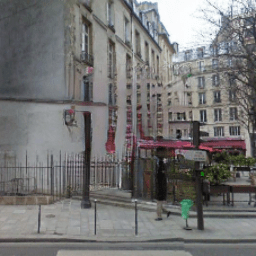}&
\includegraphics[width=0.3\linewidth]{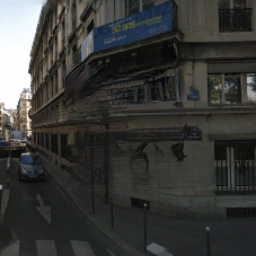}&
\includegraphics[width=0.3\linewidth]{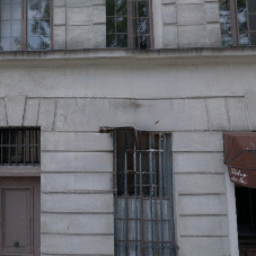}\\

\rotatebox[origin=l]{90}{PGGAN}&
\includegraphics[width=0.3\linewidth]{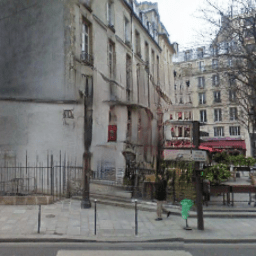}&
\includegraphics[width=0.3\linewidth]{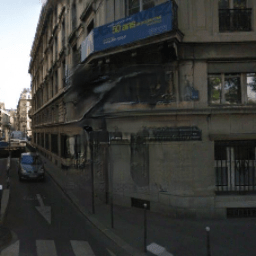}&
\includegraphics[width=0.3\linewidth]{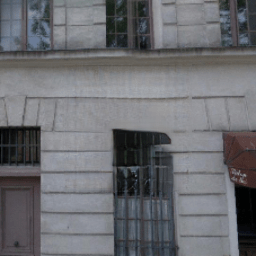}\\

\end{tabular}
   \caption{Results are obtained by training the same generator network with different discriminator architectures.}
\label{fig:gan_cmp}
\end{figure}
%-------------------------------------------------------------------------

\subsection{Ablation study}
In order to analyze effects of different components introduced, we perform several experiments by changing parameters one at a time. First, we compare the different discriminator architectures on the same generator network ResNet. All the networks are trained until no significant change occurs. Figure \ref{fig:gan_cmp} shows sample results. It can be observed for instance in the last column, the window details are reconstructed differently across the methods. As expected, the G-GAN discriminator aids in completing only the coarse image  structures. PatchGAN demonstrates significant improvement compared to G-GAN but reconstructed images still have a sign of global misconception. PGGAN blends both local and global structure and provides visually more plausible results.

\begin{figure}[t]
   \centering
\begin{tabular}{c@{\hskip 0.04in}c@{\hskip 0.04in}c@{\hskip 0.04in}c}

\rotatebox[origin=l]{90}{tconv}&
\includegraphics[width=0.3\linewidth]{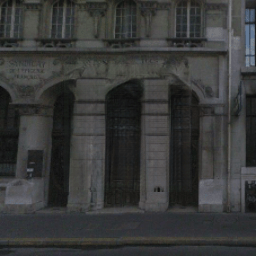}&
\includegraphics[width=0.3\linewidth]{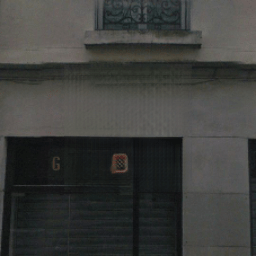}&
\includegraphics[width=0.3\linewidth]{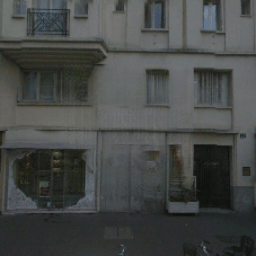}\\

\rotatebox[origin=l]{90}{iconv}&
\includegraphics[width=0.3\linewidth]{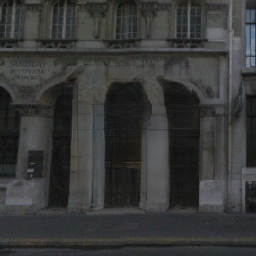}&
\includegraphics[width=0.3\linewidth]{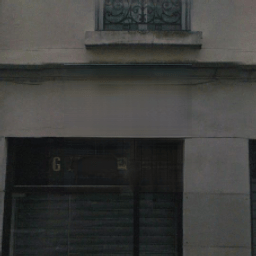}&
\includegraphics[width=0.3\linewidth]{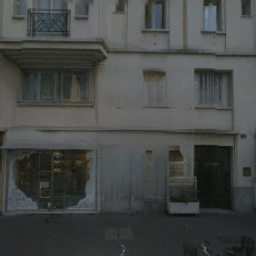}\\

\end{tabular}
   \caption{Sample outputs; top: transposed convolution (tconv) and bottom: interpolated convolution (iconv) \cite{iconv}. }
\label{fig:upsample_cmp}
\end{figure}

Along with the discriminator design, another important factor for image synthesis is the layers used in generator network models. In this study, we prefer interpolated convolution rather than transposed convolution because it provides smooth outputs. To illustrate the impact of the interpolated convolution, we tested the same PGGAN except the upsampling layer as demonstrated in Figure \ref{fig:upsample_cmp}.

%-------------------------------------------------------------------------
%-------------------------------------------------------------------------

Impact of the interpolated convolution can be clearly observed by zooming to the results of Figure \ref{fig:upsample_cmp}. It clears the noise also known as checkerboard artifacts caused by the transposed convolution. However, there are examples that have more consistent structures obtained by the transposed convolution (e.g. see the first column of the figure). These layers have distinct characteristics that each direct the generator to a different point in the solution space. Both layers should be analyzed further which is not in the scope of this study.
%-------

\begin{figure}[h]
\begin{center}
	\includegraphics[width=0.9\linewidth]{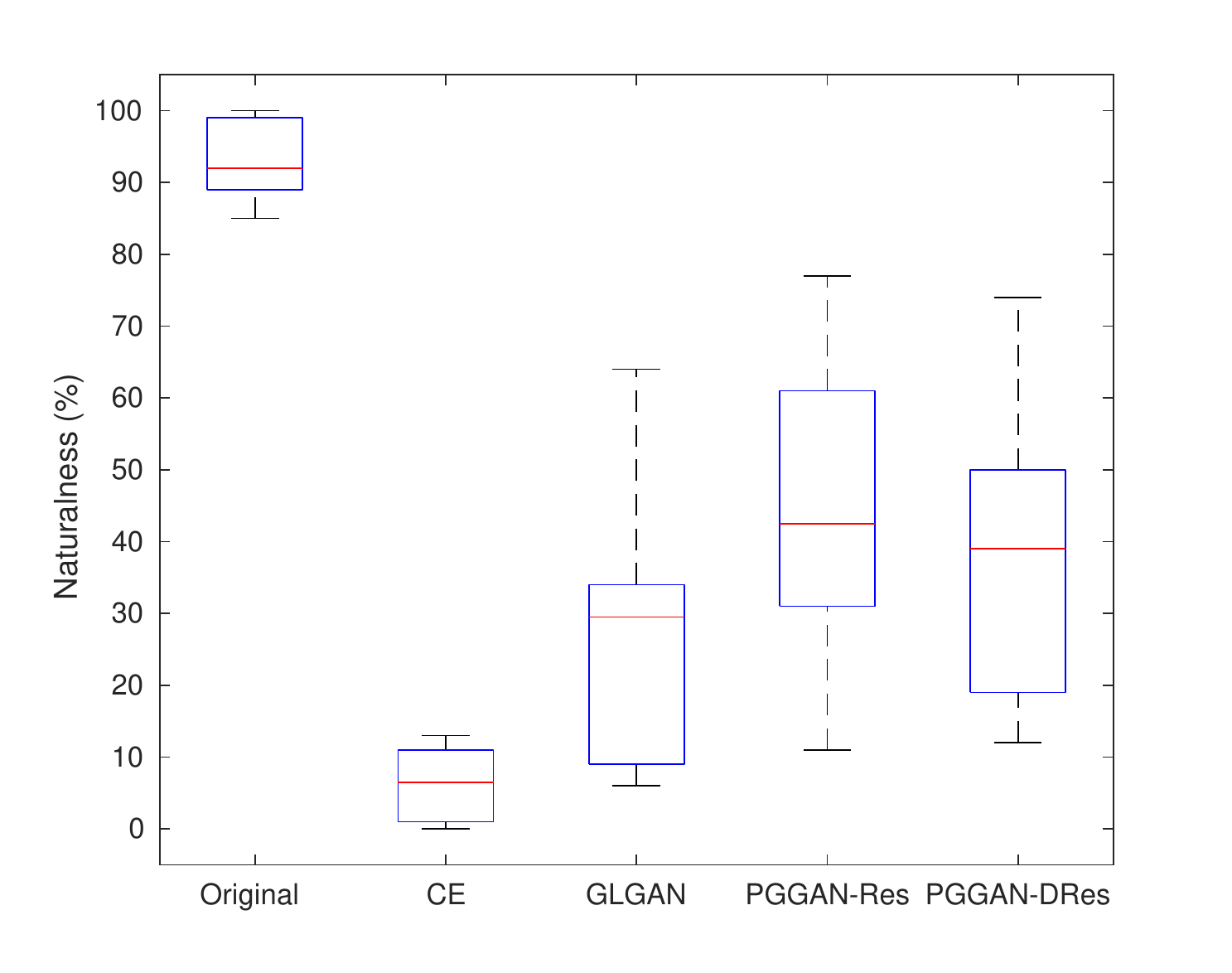}
\end{center}
   \caption{Perceptual comparison of Paris \cite{psv} images inpainted by different approaches.}
\label{fig:user_score_cmp}
\end{figure}

\subsection{Comparative evaluation}
We compare our PGGAN with ResNet (PGGAN-Res) and PGGAN with ResNet-Dilated convolution (PGGAN-DRes) to three current inpainting methods: (i) CE-Context-Encoder is adapted from \cite{context_encoder} to work with 256x256 images where full images are reconstructed; (ii) GLGAN \cite{glgan} over 256x256 images; (iii) Neural Patch Synthesis (NPS) \cite{high_res_mc} over 512x512 images. 

\textbf{Speed:} As PGGAN and GLGAN are both end-to-end texture generators, their computation times are similar on the order of miliseconds. On the other hand, NPS approach takes several seconds due to their local texture constraint.

\textbf{PSNR} and \textbf{SSIM} \cite{ssim} are the two mostly used evaluation criteria among the image generation community  although it is known that they are not sufficient for quality assessment. Nonetheless, in order to quantitatively compare our method with the current works, we report PSNR, SSIM, mean L1, and mean L2 loss in Table \ref{table:psnr_256} and Table \ref{table:psnr_512} for 256x256 and 512x512 images, respectively.

\begin{table}[h]
\begin{center}
\begin{tabular}{|l|c|c|c|c|}
\hline
Method &L1 Loss&L2 Loss&psnr(dB)&ssim \\
\hline\hline
CE \cite{context_encoder} & 6.21 & 1.34 &  18.12 & 0.838 \\
\hline
GLGAN\cite{glgan} & 5.82 & 2.33  & 18.28 & 0.863\\
\hline
PGGAN-DRes  & \textbf{5.54} & \textbf{1.19} & \textbf{19.03} & \textbf{0.866}\\
\hline
PGGAN-Res & 5.46 & 1.2 & 18.92 & 0.865\\
\hline
\end{tabular}
\end{center}
\caption{Performance comparison on 256x256 images from Paris Street View evaluation set.}
\label{table:psnr_256}
\end{table}

%-------

\begin{table}[h]
\begin{center}
\begin{tabular}{|l|c|c|c|c|}
\hline
Method &L1 Loss&L2 Loss&psnr(dB)&ssim \\
\hline\hline
NPS\cite{high_res_mc} & 10.01 & 2.21  & 18.0 & -\\
\hline
PGGAN-DRes  & \textbf{5.42} & \textbf{1.16} & \textbf{18.9} & 0.884\\
\hline
\end{tabular}
\end{center}
\caption{Comparison between NPS and our DRes-PGGAN with 512x512 Paris Street View images.}
\label{table:psnr_512}
\end{table}

PGGAN achieves an improvement in all measures for both 512x512 and 256x256 images. These results are also supported by perceptual and visual evaluations as presented next.

%-------------------------------------------------------------------------

\begin{figure*}
   \centering
\begin{tabular}{c@{\hskip 0.04in}c@{\hskip 0.04in}c@{\hskip 0.04in}c@{\hskip 0.04in}c}
Input&CE\cite{context_encoder}&GLGAN \cite{glgan}&PGGAN-DRes(Ours)&PGGAN-Res (Ours)\\
\includegraphics[width=0.18\linewidth]{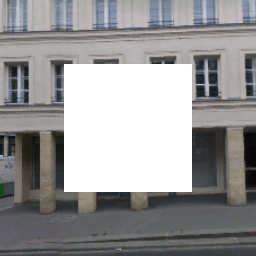}&
\includegraphics[width=0.18\linewidth]{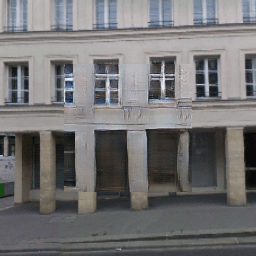}&
\includegraphics[width=0.18\linewidth]{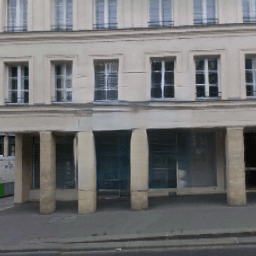}&
\includegraphics[width=0.18\linewidth]{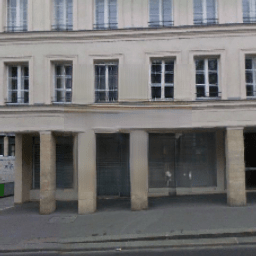}&
\includegraphics[width=0.18\linewidth]{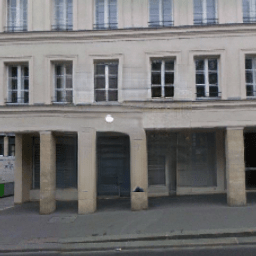}\\

\includegraphics[width=0.18\linewidth]{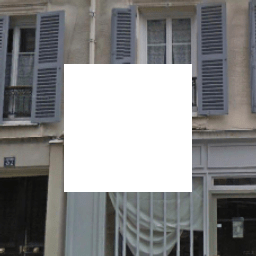}&
\includegraphics[width=0.18\linewidth]{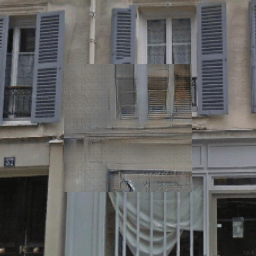}&
\includegraphics[width=0.18\linewidth]{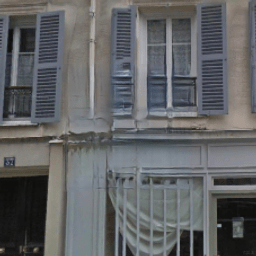}&
\includegraphics[width=0.18\linewidth]{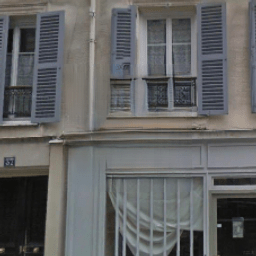}&
\includegraphics[width=0.18\linewidth]{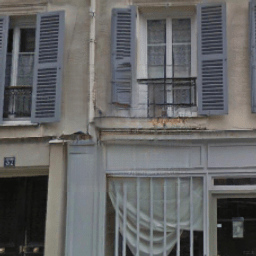}\\

\includegraphics[width=0.18\linewidth]{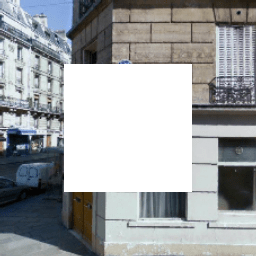}&
\includegraphics[width=0.18\linewidth]{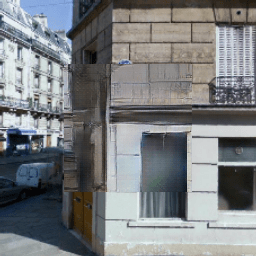}&
\includegraphics[width=0.18\linewidth]{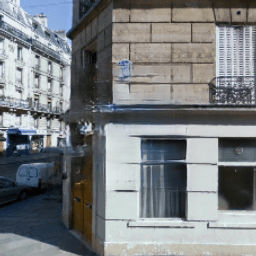}&
\includegraphics[width=0.18\linewidth]{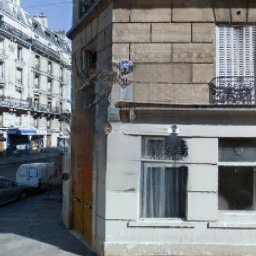}&
\includegraphics[width=0.18\linewidth]{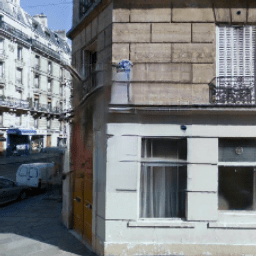}\\

\includegraphics[width=0.18\linewidth]{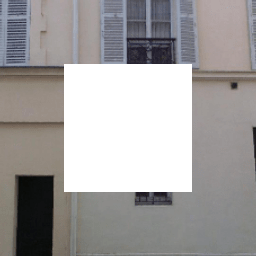}&
\includegraphics[width=0.18\linewidth]{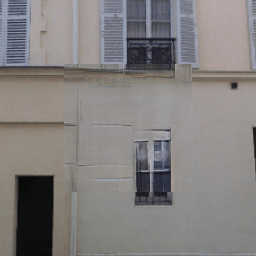}&
\includegraphics[width=0.18\linewidth]{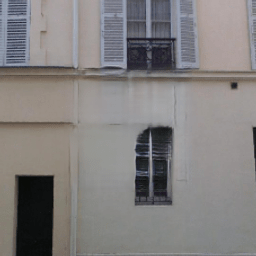}&
\includegraphics[width=0.18\linewidth]{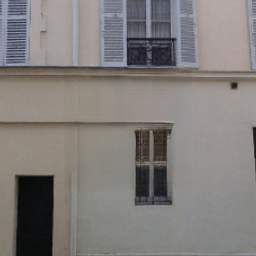}&
\includegraphics[width=0.18\linewidth]{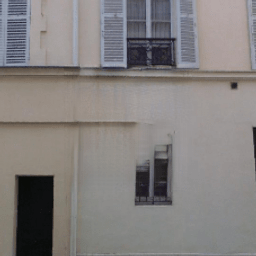}\\

\includegraphics[width=0.18\linewidth]{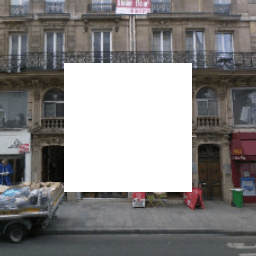}&
\includegraphics[width=0.18\linewidth]{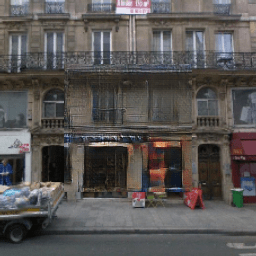}&
\includegraphics[width=0.18\linewidth]{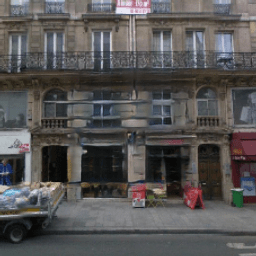}&
\includegraphics[width=0.18\linewidth]{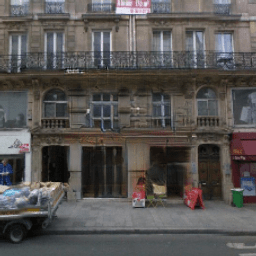}&
\includegraphics[width=0.18\linewidth]{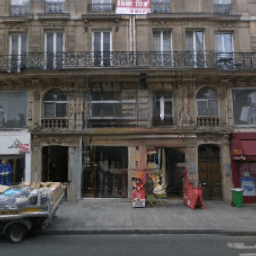}\\

\end{tabular}
    \caption{Visual comparison on 256x256 Paris Street View Dataset \cite{psv}. }
    \label{fig:psv_256_cmp} % I can do without the label too
\end{figure*}

%-------------------------------------------------------------------------

\subsection{Perceptual evaluation}
We perform perceptual evaluation among PGGAN-Res, PGGAN-DRes, CE and GLGAN. 12 voters from our laboratory scored naturalness (as natural/not natural) of the original images and inpainting results of the methods. Overall each tester evaluated randomly sorted and blinded 500 images (5 x 100 images of the Paris Street View validation set). Figure \ref{fig:user_score_cmp} shows the boxplot of the percent naturalness score accumulated over users for each method. 

Results indicate that CE presented for 128x128 images has low performance on the 256x256 test images as also reported in \cite{context_encoder}. Rest of the methods performed similarly however, slightly better scores for PGGAN were obtained. This suggests that further emphasis of local coherence along with global structure can help to generate more plausible textures.

%-------
\subsection{Visual results}
\begin{figure*}
   \centering
\begin{tabular}{lc@{\hskip 0.04in}c@{\hskip 0.04in}c@{\hskip 0.04in}c@{\hskip 0.04in}c}

\rotatebox[origin=l]{90}{Input}&
\includegraphics[width=0.18\linewidth]{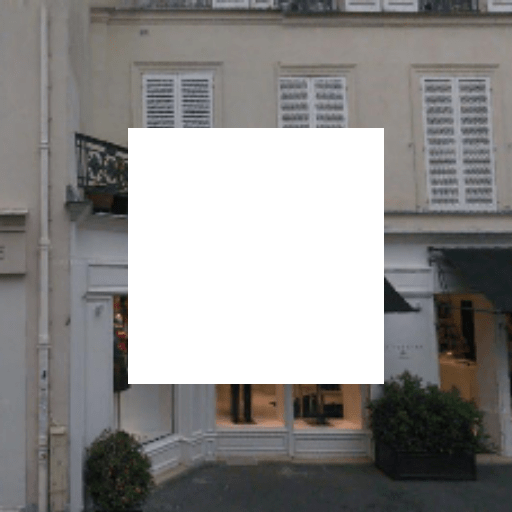}&
\includegraphics[width=0.18\linewidth]{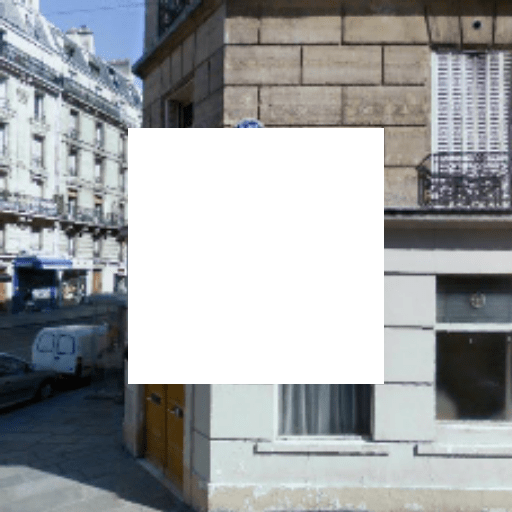}&
\includegraphics[width=0.18\linewidth]{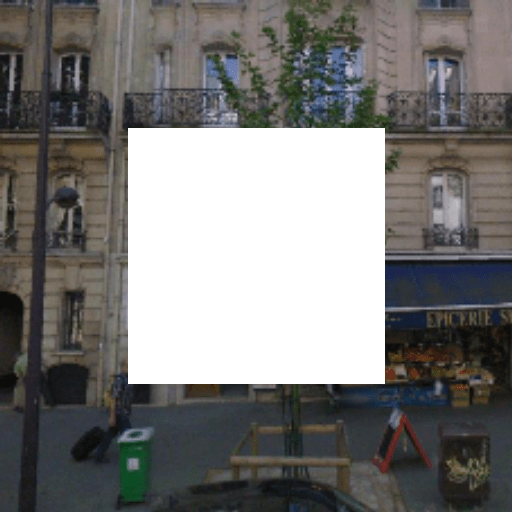}&
\includegraphics[width=0.18\linewidth]{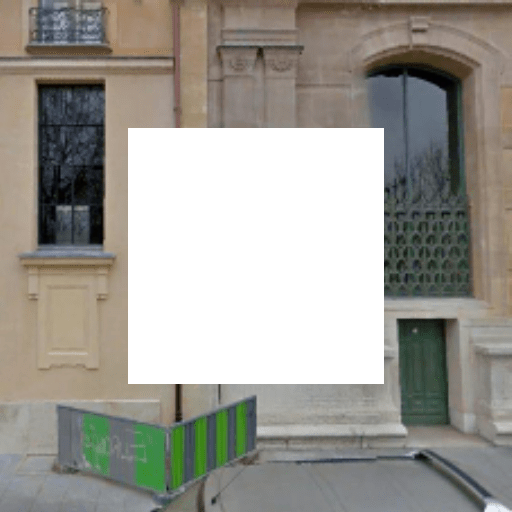}&
\includegraphics[width=0.18\linewidth]{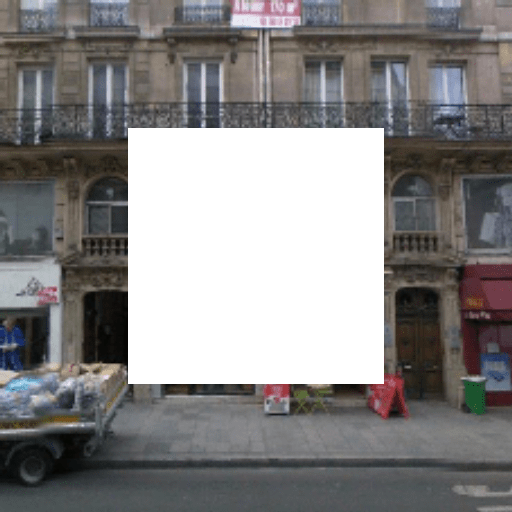}\\

\rotatebox[origin=l]{90}{NPS \cite{high_res_mc}}&
\includegraphics[width=0.18\linewidth]{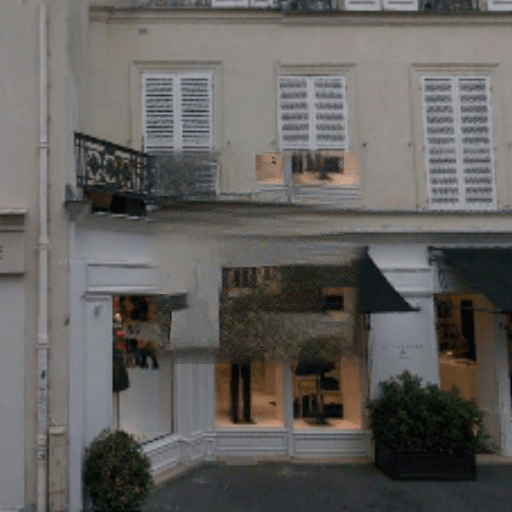}&
\includegraphics[width=0.18\linewidth]{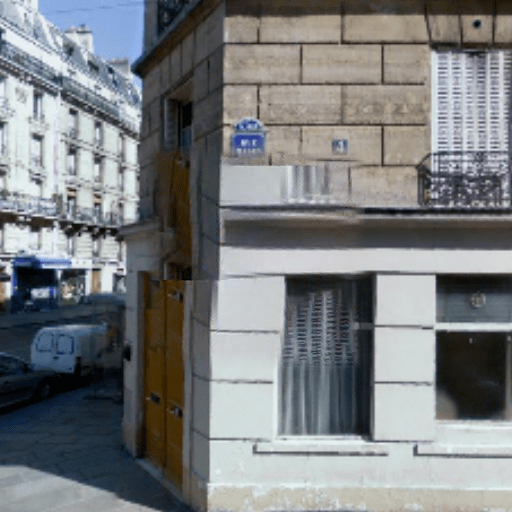}&
\includegraphics[width=0.18\linewidth]{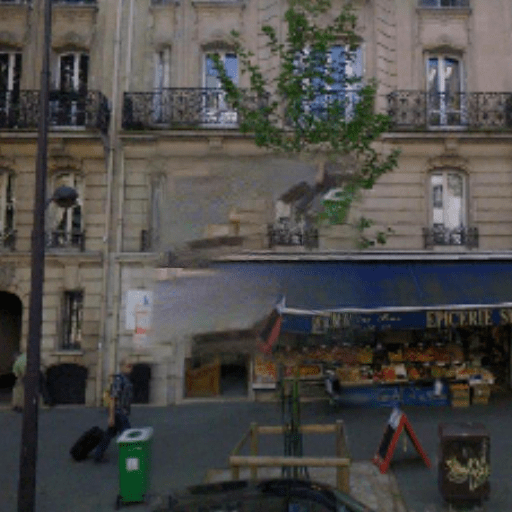}&
\includegraphics[width=0.18\linewidth]{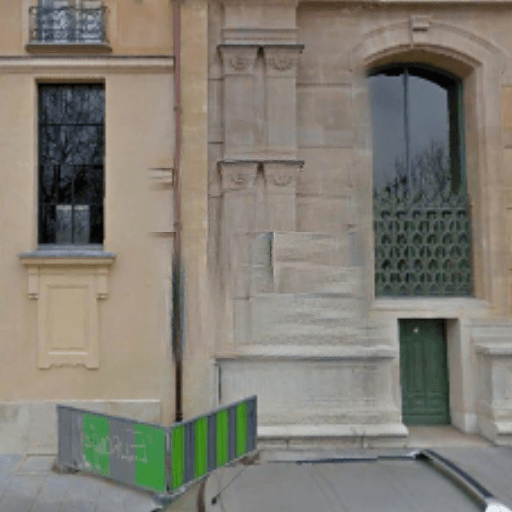}&
\includegraphics[width=0.18\linewidth]{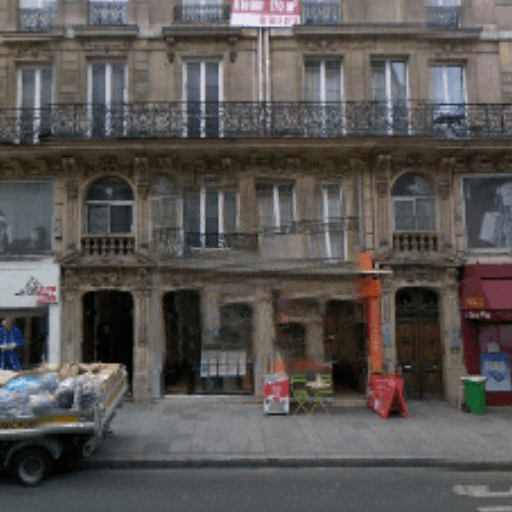}\\

\rotatebox[origin=l]{90}{PGGAN-RES}&
\includegraphics[width=0.18\linewidth]{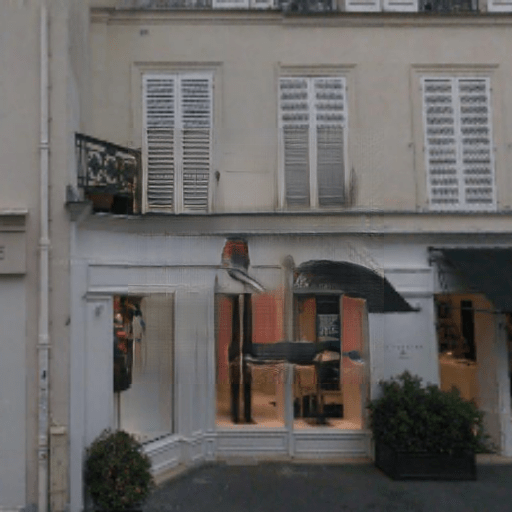}&
\includegraphics[width=0.18\linewidth]{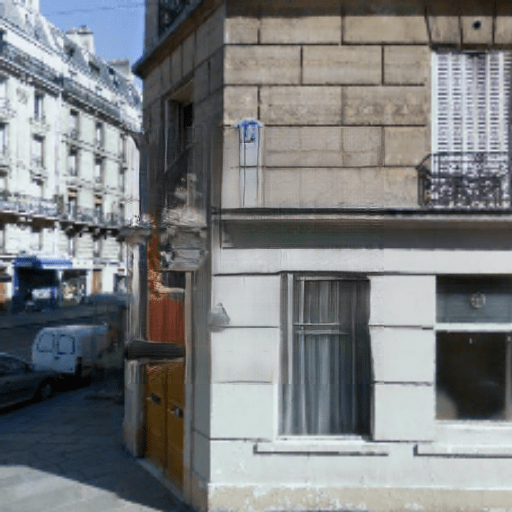}&
\includegraphics[width=0.18\linewidth]{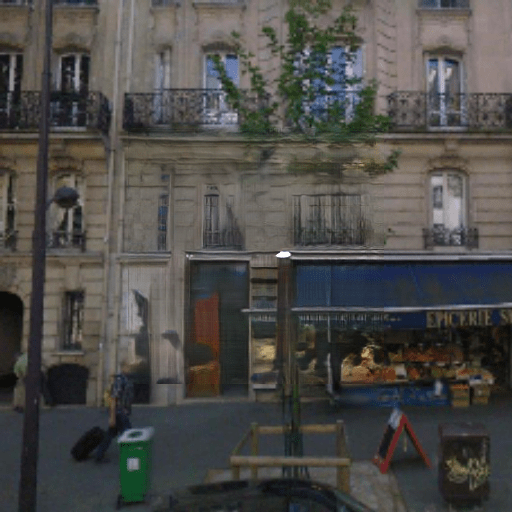}&
\includegraphics[width=0.18\linewidth]{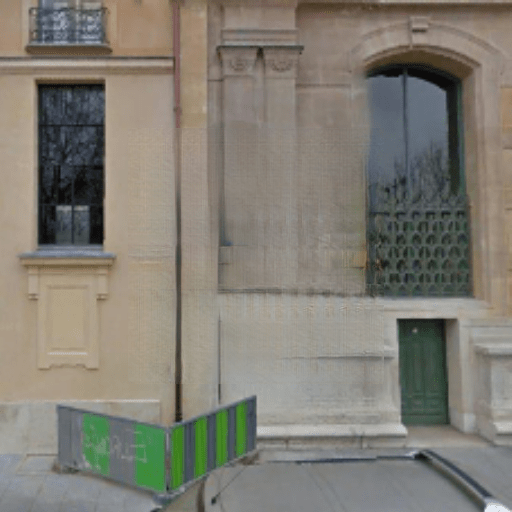}&
\includegraphics[width=0.18\linewidth]{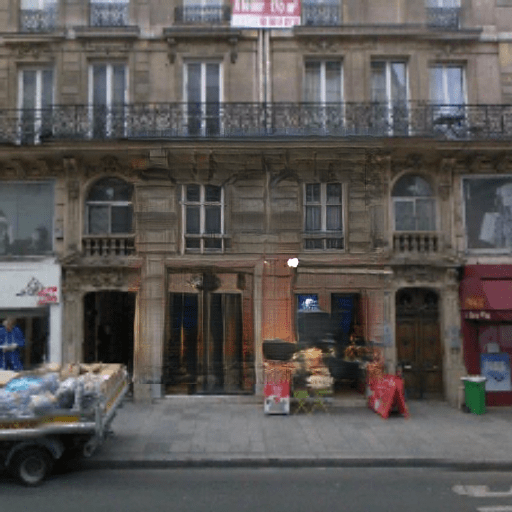}\\

\end{tabular}
    \caption{Visual comparison between PGGAN-RES and NPS \cite{high_res_mc} on 512x512 Paris Street View Dataset \cite{psv}. }
    \label{fig:psv_512_cmp} % I can do without the label too
\end{figure*}

\begin{figure}[h]
   \centering
\begin{tabular}{c@{\hskip 0.04in}c@{\hskip 0.04in}c}
\includegraphics[width=0.3\linewidth]{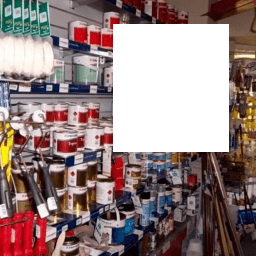}&
\includegraphics[width=0.3\linewidth]{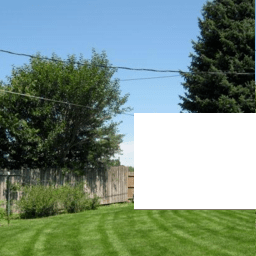}&
\includegraphics[width=0.3\linewidth]{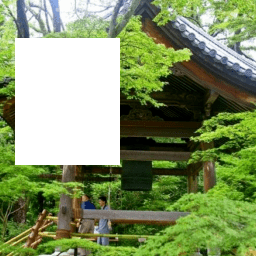}\\

\includegraphics[width=0.3\linewidth]{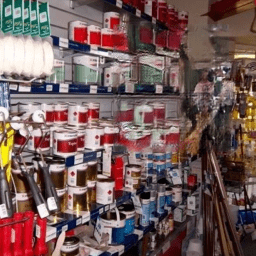}&
\includegraphics[width=0.3\linewidth]{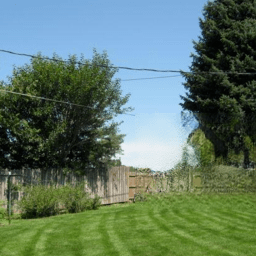}&
\includegraphics[width=0.3\linewidth]{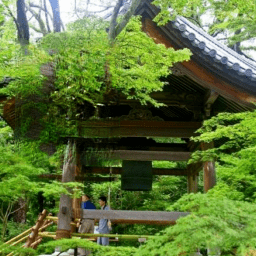}\\

\end{tabular}
   \caption{Sample outputs of PGGAN-DRes on Places dataset \cite{places}. }
\label{fig:places_sample}
\end{figure}

\begin{figure}[h]
   \centering
\begin{tabular}{c@{\hskip 0.04in}c@{\hskip 0.04in}c}
\includegraphics[width=0.3\linewidth]{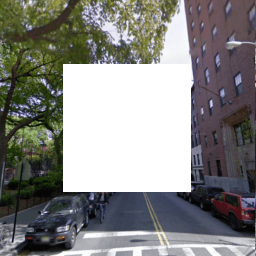}&
\includegraphics[width=0.3\linewidth]{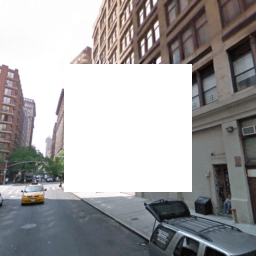}&
\includegraphics[width=0.3\linewidth]{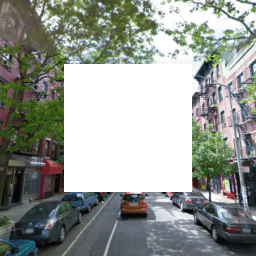}\\

\includegraphics[width=0.3\linewidth]{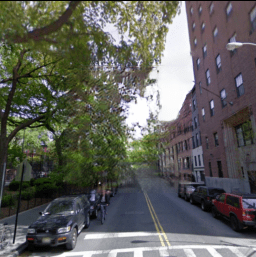}&
\includegraphics[width=0.3\linewidth]{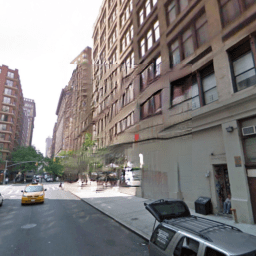}&
\includegraphics[width=0.3\linewidth]{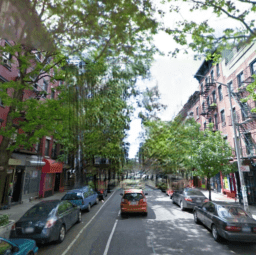}\\

\end{tabular}
   \caption{Sample outputs of PGGAN-DRes on Google Street View dataset \cite{crcv}. }
\label{fig:gsv_sample}
\end{figure}

We compare visual performance of PGGAN, NPS, and GLGAN on the common Paris Street View dataset. Figures \ref{fig:psv_256_cmp} and \ref{fig:psv_512_cmp} show the results for images of size 256x256 and 512x512 respectively. Some fail case results can be seen in Figure \ref{fig:non_cherry}. Results from Places and Google Street View datasets\footnote{See supplementary materials for extensive results.} are shown in Figures \ref{fig:places_sample} and \ref{fig:gsv_sample}.

%Dilated resnet version of PGGAN performs better in synthesis of tile texture as it can sample from the farther regions in the image without downsampling (e.g. see Figure XX row x
%-------------------------------------------------------------------------

%-------------------------------------------------------------------------
\begin{figure*}
   \centering
\begin{tabular}{c@{\hskip 0.04in}c@{\hskip 0.04in}c@{\hskip 0.04in}c@{\hskip 0.04in}c}
\includegraphics[width=0.18\linewidth]{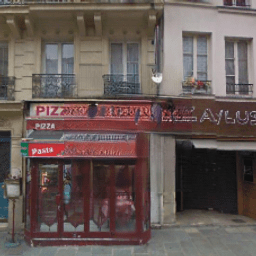}&
\includegraphics[width=0.18\linewidth]{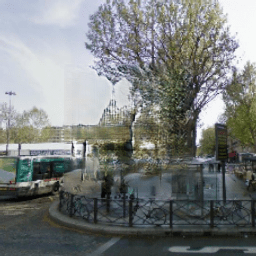}&
\includegraphics[width=0.18\linewidth]{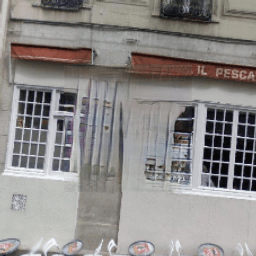}&
\includegraphics[width=0.18\linewidth]{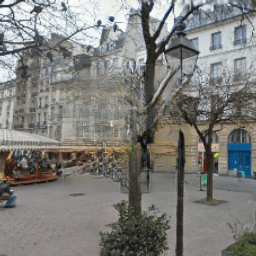}&
\includegraphics[width=0.18\linewidth]{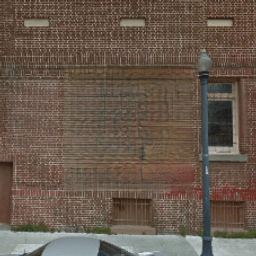}\\

\end{tabular}
    \caption{Non-cherry picked results from PGGAN-DRes. }
    \label{fig:non_cherry} % I can do without the label too
\end{figure*}

%-------------------------------------------------------------------------

%-------------------------------------------------------------------------
%-------------------------------------------------------------------------

%-------------------------------------------------------------------------
%-------------------------------------------------------------------------

%-------------------------------------------------------------------------

\section{Conclusion}
\label{sec:conc}
The image inpainting results in this paper suggest that low-level merging then high-level splitting a patch-based technique such as PatchGAN with a traditional GAN network can aid in acquiring local continuity of image texture while conforming to the holistic nature of the images. This merger produces visually and quantitatively better results than the current inpainting methods. However, the  inpainting problem which is tightly coupled to the generative modeling problem is still open to further progress.

%-------------------------------------------------------------------------
%\clearpage
\newpage

{\small
\bibliographystyle{ieee}
\bibliography{egbib}
}

\clearpage
\newpage

\includepdf[pages=1]{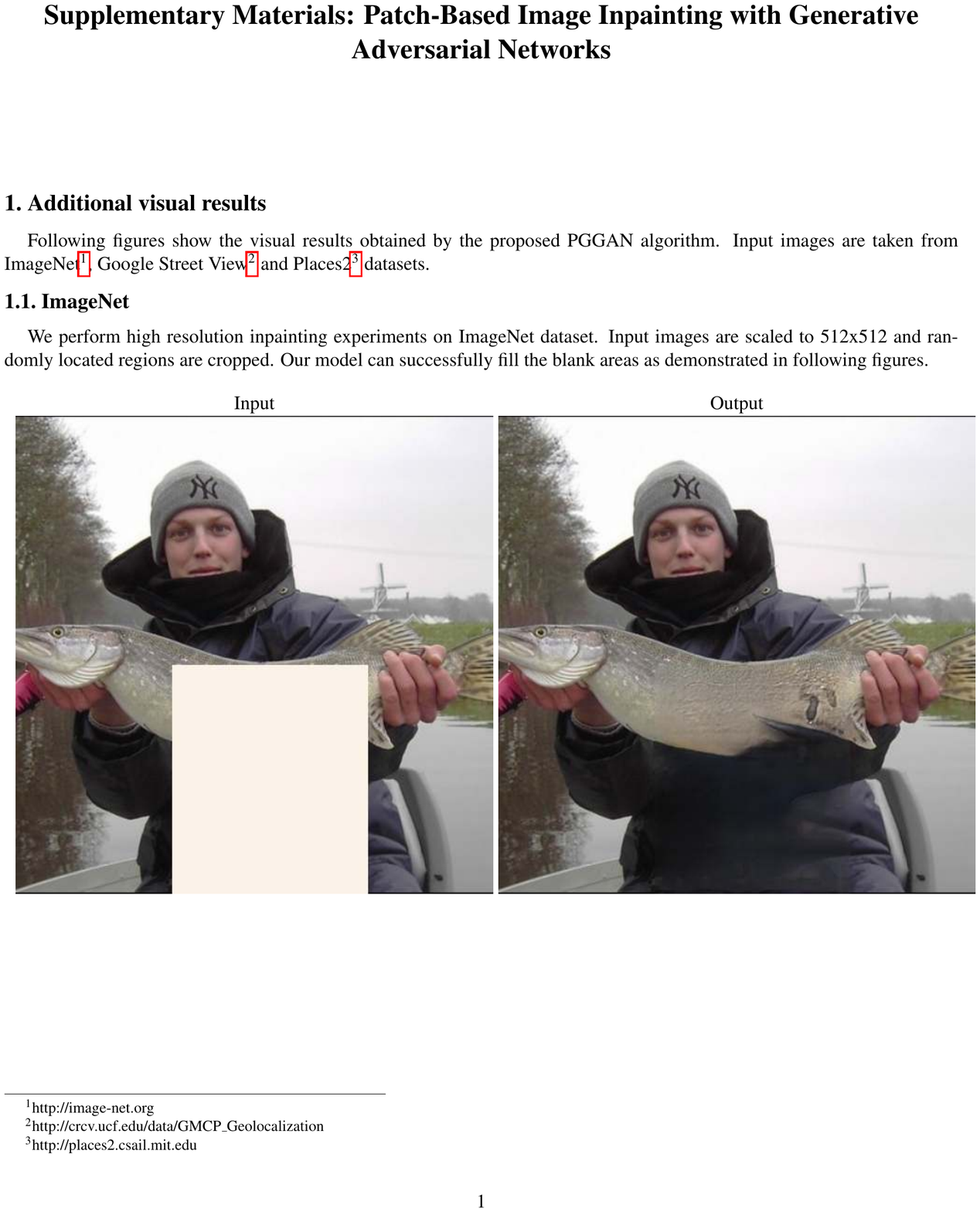}
\includepdf[pages=2]{supp.pdf}
\includepdf[pages=3]{supp.pdf}
\includepdf[pages=4]{supp.pdf}
\includepdf[pages=5]{supp.pdf}
\includepdf[pages=6]{supp.pdf}
\includepdf[pages=7]{supp.pdf}
\includepdf[pages=8]{supp.pdf}
\includepdf[pages=9]{supp.pdf}
\includepdf[pages=10]{supp.pdf}
\includepdf[pages=11]{supp.pdf}
\includepdf[pages=12]{supp.pdf}
\includepdf[pages=13]{supp.pdf}
\includepdf[pages=14]{supp.pdf}
\includepdf[pages=15]{supp.pdf}
\includepdf[pages=16]{supp.pdf}
\includepdf[pages=17]{supp.pdf}
\includepdf[pages=18]{supp.pdf}

\end{document}